\documentclass{article}

\usepackage[dblblindworkshop, final]{neurips_2026}
\workshoptitle{PTA: From Pretrained Representations to Acting Agents}

\usepackage[utf8]{inputenc} 
\usepackage[T1]{fontenc}    
\usepackage{hyperref}       
\usepackage{url}            
\usepackage{booktabs}       
\usepackage{amsfonts}       
\usepackage{nicefrac}       
\usepackage{microtype}      
\usepackage{xcolor}         
\usepackage{amsmath}
\usepackage{graphicx}
\usepackage{wrapfig}

\title{When Does Retrieval Help? A Study of In-Context Adaptation in Vision-Language-Action Models}

\author{%
  Zixuan Liu$^{1}$\thanks{Equal contribution.} \quad
  Joris Köster$^{2}$\footnotemark[1] \quad
  Zizhan Zheng$^{1}$ \quad
  Siavash Khajavi$^{3}$ \\
  $^{1}$Department of Computer Science, Tulane University \\
  $^{2}$Department of Computer Science, Aalto University \\
  $^{3}$Department of Industrial Engineering and Management, Aalto University \\
  \texttt{\{zliu41, zzheng3\}@tulane.edu}, \quad
  \texttt{\{joris.koster, siavash.khajavi\}@aalto.fi}
}

\begin{document}

\maketitle

\begin{abstract}
Vision-language-action (VLA) models have shown strong potential as generalist robot policies, but adapting them to unseen tasks often requires costly parameter updates. Recent work such as RICL introduces in-context adaptability by retrieving expert demonstrations based on the current VLA observation and providing them as additional context at test time. The effectiveness of this adaptation therefore depends critically on the retrieval mechanism. In this work, we systematically study how different retrieval methods affect both retrieval quality and task performance within the RICL framework. Specifically, we compare four different methods: image-based retrieval, retrieval augmented with VLA's state, retrieval using features from the VLA backbone, and random retrieval. Our experiments yield three main findings. First, no retrieval method consistently dominates the others in task success, while surprisingly, random retrieval achieves a non-trivial success rate. Second, standard retrieval-quality diagnostics do not reliably reflect downstream VLA performance. Third, demonstrations from different but related tasks can provide useful transferable information. Together, these results provide an initial step toward understanding how retrieval mechanisms shape the in-context learning capability of VLA models and their downstream task performance, while highlighting the need for more careful design and evaluation of retrieval mechanisms for reliable test-time adaptation.
\end{abstract}

\section{Introduction}

Recent advances in vision-language-action (VLA) models~\cite{black2024pi_0,kim2024openvla,pertsch2025fast,team2024octo,bousmalis2023robocat,brohan2023rt,vuong2023open,team2025gemini} have enabled a single policy to perform a broad range of robotic manipulation tasks, demonstrating their potential as general-purpose robotic control policies. However, even strong pre-trained VLAs often require substantial fine-tuning or retraining to perform reliably under distribution shifts, such as new environments, unseen objects, or novel task instructions~\cite{zhang2026retrieval}, which limits rapid adaptation at deployment. In contrast, large language models (LLMs)~\cite{achiam2023gpt,anthropic2024claude,team2023gemini,liu2024enhancing,liu2026targeting,liu2026ua,liu2025llava,liu2026risk,koster2026memboost} can often adapt to new tasks at inference time through in-context learning (ICL): given only a few input-output examples as context, an LLM can adapt to a new task without updating its parameters~\cite{brown2020language,dong2024survey,mao2025survey}. Inspired by this paradigm, RICL~\cite{sridhar2025ricl} equips a pre-trained VLA with in-context adaptability by conditioning action prediction on previously collected expert demonstrations, enabling adaptation to unseen tasks without test-time parameter updates. Specifically, RICL maintains a small retrieval buffer of expert demonstrations and, given the current observation, retrieves relevant demonstrations as additional context for the VLA policy. The effectiveness of such retrieval-based in-context adaptation therefore depends critically on which demonstrations are selected from the retrieval buffer, since the retrieved examples determine what task-specific information is provided to the VLA at test time. 

In this work, we study how the current observation should be represented for retrieval so that the returned demonstrations are most useful for adapting to an unseen task. The original RICL framework performs retrieval using only the embedding of the current image. We investigate whether incorporating additional information, such as the robot's proprioceptive state or features extracted directly from the VLA backbone, can lead to more effective retrieval and improved task performance. Our main findings are: 1. \textbf{No retrieval representation clearly dominates.} Surprisingly, even random retrieval achieves a non-trivial success rate, suggesting that the VLA can still benefit from demonstrations even when they are not selected based on similarity to the current observation. 2. \textbf{Better retrieval quality does not necessarily translate into better downstream control performance.} 3. \textbf{Retrieval from different but related tasks can provide useful transferable information.} Together, these results suggest that improving retrieval for VLA adaptation requires going beyond simple similarity matching between the current observation and stored demonstrations, as well as straightforward extensions such as directly incorporating additional information. More importantly, retrieval-quality metrics should be designed to better reflect their effectiveness during actual VLA execution.

\vspace{-1ex}

\section{Preliminaries: RICL Framework}

\vspace{-1ex}

In this paper, we consider a general robotic manipulation setting following prior work~\cite{zhang2026retrieval,sridhar2025ricl}, where a pre-trained vision-language-action (VLA) policy \(\pi_\theta\) is required to perform novel tasks in unseen environments without any test-time parameter updates. Each task is specified by a natural language instruction \(x^{\mathrm{text}}\). At timestep \(t\), the agent observes multi-view images \(I_t=(I_t^{\mathrm{top}},I_t^{\mathrm{side}},I_t^{\mathrm{wrist}})\), corresponding to the top, side, and wrist camera views, together with a proprioceptive state \(s_t\). A standard VLA policy predicts an action according to \(\hat a_t \sim \pi_\theta(\cdot \mid I_t,s_t,x^{\mathrm{text}})\).

Although pre-trained VLA policies have shown promising generalist robot-control capabilities, their performance often remains unreliable under distribution shifts~\cite{black2024pi_0,kim2024openvla}.~\cite{sridhar2025ricl} propose Re-training for In-Context Learning (RICL), which enables a VLA policy to use retrieved expert demonstrations as in-context examples to improve test-time adaptability. Let \(\mathcal D=\{m_i\}_{i=1}^{N}\) denote a small retrieval buffer consisting of \(N\) expert demonstration frames, where each entry is \(m_i=(I_i,s_i,x_i^{\mathrm{text}},a_i)\). Here, \(I_i=(I_i^{\mathrm{top}},I_i^{\mathrm{side}},I_i^{\mathrm{wrist}})\) denotes the multi-view images at the demonstration state, \(s_i\) is the corresponding proprioceptive state, \(x_i^{\mathrm{text}}\) is the task instruction, and \(a_i\) is the expert action or action chunk (i.e., an array of actions over many timesteps). These demonstrations are not used to update the VLA parameters at test time. Instead, they serve as external context for adapting the policy to the current task. At each timestep \(t\), RICL retrieves a small set of relevant entries from the demonstration buffer \(\mathcal D\). Let \(f_{\mathrm{DINO}}\) denote a frozen DINOv2 image encoder~\cite{oquab2023dinov2}. RICL uses the top-view image \(I_t^{\mathrm{top}}\) as the retrieval query and compares it with the top-view image of each demonstration entry. Specifically, for each entry \(m_i\in\mathcal D\), it computes the distance \(d_t(i)=\|f_{\mathrm{DINO}}(I_t^{\mathrm{top}})-f_{\mathrm{DINO}}(I_i^{\mathrm{top}})\|_2\). The \(M\) nearest entries are then selected by \((j_1,\ldots,j_M)=\operatorname{TopM}_{i\in\{1,\ldots,N\}}^{\mathrm{smallest}} d_t(i)\), with \(d_t(j_1)\le d_t(j_2)\le \cdots \le d_t(j_M)\). The retrieved context is \(\mathcal M_t=(m_{j_1},\ldots,m_{j_M})\), where entries are ordered from the closest to the farthest retrieved neighbor. The RICL policy then conditions on both the retrieved context and the current state to predict the next action: \(\hat a_t \sim \pi_\theta^{\mathrm{RICL}}(\cdot \mid \mathcal M_t,I_t,s_t,x^{\mathrm{text}})\). Thus, RICL converts the original VLA action prediction problem into a retrieval-conditioned action prediction problem, where the policy is expected to infer the appropriate action by comparing the current observation with a small set of retrieved expert examples. We include a detailed related work on in-context learning for VLA models in Appendix~\ref{app:related-work}.

\vspace{-1ex}
\section{Empirical Study of Retrieval for VLA Adaptation}
\vspace{-1ex}
In this section, we conduct an empirical study to answer the following research questions: (1) Which retrieval representation works best? (2) Do standard retrieval-quality diagnostics reliably predict downstream control performance? (3) Is retrieval from a different task necessarily harmful?

\subsection{Experimental Setup}
For all experiments, we mainly follow the setup in RICL~\cite{sridhar2025ricl}, with important details and several changes stated below. Additional implementation details are provided in Appendix~\ref{app:implement}.

\textbf{Evaluation Benchmark.}
We evaluate all methods on the \textbf{LIBERO} benchmark~\cite{liu2023libero}, which was not considered in the original RICL study. LIBERO is a simulated robotic manipulation benchmark built around the Franka Panda arm. For retrieval, we construct a shared demonstration buffer containing 20 expert demonstrations per task, and all retrieval methods operate on this same buffer. For evaluation, we select 25 evaluation tasks, and each task is tested on 10 fixed initial scenes that are disjoint from the demonstration scenes used to construct the retrieval buffer.

\textbf{Model Architecture.}
We evaluate all retrieval methods using a fixed \(\pi_0\)-FAST VLA model~\cite{pertsch2025fast}, following the RICL setting. Since the original RICL model is trained for a different benchmark, we further train the model on LIBERO using a comparable RICL post-training procedure. 
 
\textbf{Retrieval Methods.}
We compare four retrieval methods. \texttt{RICL-Image} follows the original RICL setting and retrieves demonstrations using only the DINOv2 embedding of the top-camera image. \texttt{RICL-ImageState} augments image-based retrieval with the proprioceptive state \(s_t\), allowing the retrieval score to account for both visual and robot-state similarity. We include this method to evaluate whether incorporating state information improves retrieval. \texttt{RICL-VLAFeat} performs retrieval using features extracted directly from the \(\pi_0\)-FAST model, taking the top-camera image and proprioceptive state as inputs. This method tests whether representations learned by the VLA itself provide a more effective retrieval than generic visual features. \texttt{RICL-Random} samples entries uniformly from the demonstration buffer and serves as a baseline for comparison. All methods retrieve \(M=4\) entries from the same demonstration buffer.

\subsection{Results}

\begin{table}[t]
\centering
\caption{Main results for the four retrieval methods and paired comparisons. \emph{\(\Delta\)} denotes the difference between two methods. A smaller McNemar-\(p\) provides stronger evidence that the two methods differ significantly in success rate. }
\label{tab:main-results}
\scriptsize
\setlength{\tabcolsep}{2.2pt}
\begin{minipage}[t]{0.41\linewidth}
\centering
\textbf{(a) Main result}

\begin{tabular}{lcc}
\toprule
\textbf{Method} & \textbf{Succ.} & \textbf{Task-hier. CI} \\
\midrule
\texttt{RICL-Image}      & 40.0 & [27.2, 54.0] \\
\texttt{RICL-ImageState} & 36.8 & [24.0, 50.0] \\
\texttt{RICL-VLAFeat}    & 40.4 & [28.4, 52.8] \\
\texttt{RICL-Random}     & 26.0 & [15.6, 37.6] \\
\bottomrule
\end{tabular}
\end{minipage}
\hfill
\begin{minipage}[t]{0.57\linewidth}
\centering
\textbf{(b) Paired comparisons}

\begin{tabular}{lccc}
\toprule
\textbf{Comparison} 
& \(\boldsymbol{\Delta}\) \textbf{Succ.} 
& \(\boldsymbol{\Delta}\) \textbf{Task-hier. CI} 
& \textbf{McNemar \(p\)} \\
\midrule
\texttt{RICL-ImageState}, \texttt{RICL-Image}
    & -3.2 & [-13.2, 6.8] & 0.374 \\
\texttt{RICL-VLAFeat}, \texttt{RICL-ImageState}
    & +3.6 & [-6.0, 13.6] & 0.349 \\
\texttt{RICL-VLAFeat}, \texttt{RICL-Image}
    & +0.4 & [-10.4, 10.8] & 1.000 \\
\midrule
\texttt{RICL-Image}, \texttt{RICL-Random}
    & +14.0 & [-0.4, 28.8] & 0.000425 \\
\texttt{RICL-ImageState}, \texttt{RICL-Random}
    & +10.8 & [-1.6, 23.6] & 0.003 \\
\texttt{RICL-VLAFeat}, \texttt{RICL-Random}
    & +14.4 & [2.8, 26.0] & 0.000044 \\
\bottomrule
\end{tabular}
\end{minipage}
\end{table}

\begin{figure}[t]
\centering
\begin{minipage}[t]{0.24\linewidth}
    \centering
    \includegraphics[width=\linewidth]{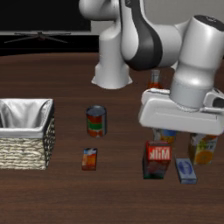}
    \vspace{-1mm}
    
    \scriptsize Task 50: pick up the alphabet soup and put it in the basket
\end{minipage}
\hfill
\begin{minipage}[t]{0.24\linewidth}
    \centering
    \includegraphics[width=\linewidth]{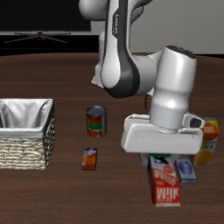}
    \vspace{-1mm}
    
    \scriptsize Task 49: pick up the tomato sauce and put it in the basket
\end{minipage}
\hfill
\begin{minipage}[t]{0.24\linewidth}
    \centering
    \includegraphics[width=\linewidth]{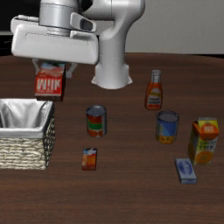}
    \vspace{-1mm}
    
    \scriptsize Task 46: pick up the alphabet soup and put it in the basket in another scene
\end{minipage}
\hfill
\begin{minipage}[t]{0.24\linewidth}
    \centering
    \includegraphics[width=\linewidth]{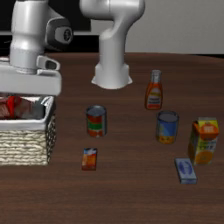}
    \vspace{-1mm}
    
    \scriptsize Task 52: pick up the milk and put it in the basket
\end{minipage}

\caption{A successful \texttt{RICL-Image} example on Task~52. The dominant retrieved source evolves from Task~50 to Task~49, Task~46, and eventually the target Task~52.}
\vspace{-2ex}
\label{fig:cross-task-example-t52}
\end{figure}

\textbf{No retrieval representation clearly dominates.} 
Table~\ref{tab:main-results} reports the main results for the four retrieval methods. In addition to the overall success rate, we report a task-hierarchical bootstrap confidence interval, denoted as \emph{task-hier. CI}. This interval captures task-level uncertainty. Since a method may achieve a high average success rate by performing well on some tasks while failing on others, task-level bootstrapping therefore provides a more conservative estimate of how reliably a retrieval method performs across different tasks. Details of the bootstrap procedure are provided in Appendix~\ref{app:implement}. We also conduct pairwise comparisons between the retrieval methods. For each comparison, we use an exact paired McNemar test~\cite{mcnemar1947note,fagerland2013mcnemar} detailed in Appendix~\ref{app:implement} and report the resulting \(p\)-value. A smaller \(p\)-value provides stronger evidence that the two methods have different success rates. Table~\ref{tab:main-results} shows that, considering success rate alone, \texttt{RICL-VLAFeat} achieves the highest performance at \(40.4\%\), slightly outperforming \texttt{RICL-Image} at \(40.0\%\) and \texttt{RICL-ImageState} at \(36.8\%\). This suggests that features extracted from the VLA backbone can provide an effective retrieval representation without relying on an additional DINOv2 encoder. Interestingly, augmenting image-based retrieval with proprioceptive state does not improve performance and instead leads to a lower numerical success rate. However, these differences are small: the pairwise comparisons among the three retrieval methods yield large McNemar-\(p\) values. Thus, none of the retrieval methods can be shown to statistically outperform the others. Another notable observation is the relatively strong performance of \texttt{RICL-Random}, which achieves a \(26.0\%\) success rate despite retrieving demonstrations uniformly at random. This suggests that VLA can still benefit from retrieved demonstrations even when they are not selected based on similarity to the current query, or that its behavior remains partially robust to irrelevant retrieved context. Although all three retrieval methods achieve higher average success than \texttt{RICL-Random}, only \texttt{RICL-VLAFeat} shows a reliably positive paired improvement, with \(\Delta\) Succ.\(=14.4\) percentage points, \(\Delta\) Task-hier. CI \(=[2.8,26.0]\), and McNemar-\(p=0.000044\). In contrast, the corresponding confidence intervals for \texttt{RICL-Image} and \texttt{RICL-ImageState} include zero, indicating that their improvements over \texttt{RICL-Random} are not consistently positive across tasks. These results suggest that VLA-backbone features may provide a more consistently useful retrieval signal relative to random retrieval under our evaluation setting, although they do not establish a clear advantage over the other retrieval methods.

\begin{wraptable}{r}{0.50\linewidth}
\vspace{-6mm}
\centering
\caption{Retrieval-quality analysis. A higher NDCG value indicates that the retrieved demonstrations are more relevant to the query.}
\label{tab:retrieval-quality}
\scriptsize
\setlength{\tabcolsep}{8pt}
\renewcommand{\arraystretch}{0.95}

\begin{tabular}{lccc}
\toprule
& \multicolumn{2}{c}{\textbf{Exact-task retrieval rate}} 
& \textbf{NDCG} \\
\cmidrule(lr){2-3}
\textbf{Method} 
& \textbf{Offline} 
& \textbf{On-Policy} 
&  \\
\midrule
\texttt{RICL-Image}      & 87.2\% & 51.8\% & 0.7461 \\
\texttt{RICL-ImageState} & 87.9\% & 48.9\% & 0.7897 \\
\texttt{RICL-VLAFeat}    & 83.1\% & 38.7\% & 0.7280 \\
\texttt{RICL-Random}     & 4.0\%  & 4.1\%  & 0.0199 \\
\bottomrule
\end{tabular}
\vspace{-2mm}
\end{wraptable}

\textbf{Retrieval quality does not reliably predict control performance.} To understand how retrieval quality relates to the actual behavior of the VLA, we consider two complementary measurements, both based on the assumption that demonstrations from the same task are more likely to provide directly relevant information. We first measure the exact-task retrieval rate, defined as the fraction of retrieved entries whose task identity matches that of the query. We evaluate this metric in two settings: an offline setting, where queries are expert states from the same evaluation scenes, and an on-policy setting, where queries are the states actually visited by the VLA during execution. This comparison reveals how retrieval behavior changes when moving from successful expert trajectories to the states encountered by the policy itself. We additionally use Normalized Discounted Cumulative Gain (NDCG)~\cite{10.1145/582415.582418} to evaluate retrieval relevance in the offline setting, where a retrieved frame is considered more relevant if it comes from the same task and its subsequent action chunk is more similar to that of the query state. Additional implementation details are provided in Appendix~\ref{app:implement}. Table~\ref{tab:retrieval-quality} shows a substantial gap between offline and on-policy exact-task retrieval. When queried offline using expert frames, all three retrieval methods return same-task demonstrations in more than \(80\%\) of the retrieved entries. However, this rate drops markedly when the query comes from states actually visited by the VLA during execution. This suggests that retrieval behavior measured offline does not directly transfer to the states encountered by the policy during execution. A similar mismatch appears when comparing offline NDCG with task success. \texttt{RICL-ImageState} achieves the highest offline NDCG of \(0.7897\), followed by \texttt{RICL-Image} at \(0.7461\) and \texttt{RICL-VLAFeat} at \(0.7280\). However, as shown in Table~\ref{tab:main-results}, \texttt{RICL-ImageState} has the lowest success rate among the three retrieval methods, while \texttt{RICL-VLAFeat} achieves the highest numerical success rate. These results indicate that better retrieval quality does not necessarily translate into better downstream control performance.

\textbf{Retrieval from different tasks can provide useful information.}
We further examine representative retrieval traces to understand whether retrieving demonstrations from a different task necessarily indicates retrieval failure. As shown in Figure~\ref{fig:cross-task-example-t52}, in one successful example on Task~52 for \texttt{RICL-Image}, the dominant retrieved source shifts across Task~50, Task~49, Task~46, and eventually the same Task~52 during task execution. This example shows that the VLA can exploit demonstrations from multiple related tasks that share transferable manipulation structure, rather than relying exclusively on exact-task retrieval. Additional examples can be found in Appendix~\ref{app:result}. In contrast, we also observe systematic retrieval between tasks with similar scene structure but different goals, such as Task~24 (put the black bowl in the bottom drawer of the cabinet) and Task~25 (put the black bowl on top of the cabinet), as shown in Appendix~\ref{app:result}, Figure~\ref{fig:t24-t25-comparison}. Although these two tasks involve the same object and environment, they require different target locations. This suggests that retrieval from a different task should not be automatically regarded as incorrect. Its usefulness depends on whether the retrieved demonstration provides action-relevant information consistent with the current task goal.

{
\small
\bibliographystyle{unsrt}
\bibliography{neurips_2026}
}


\appendix
\clearpage



\section{Conclusion}

In this work, we study how different retrieval mechanisms affect in-context adaptation for vision-language-action models within the RICL framework. Our results show that no retrieval representation consistently dominates the others in downstream task success, while random retrieval still achieves a non-trivial success rate. We further find that standard retrieval-quality diagnostics do not reliably predict downstream control performance, as retrieval behavior changes substantially when moving from expert states to the states actually encountered by the VLA. Finally, our qualitative analysis shows that demonstrations from different but related tasks can still provide useful transferable information. Together, these findings suggest that effective retrieval for VLA adaptation requires more than simple similarity matching or straightforwardly adding additional observation information. Future retrieval mechanisms should instead better account for whether a retrieved demonstration is actually useful for the current control decision, and retrieval-quality metrics should more faithfully reflect the behavior of the VLA during actual execution.

\section{Acknowledge}
We acknowledge the computational resources provided by the Aalto Science-IT project.

\section{Related Work}
\label{app:related-work}

While recent vision-language-action (VLA) models, such as OpenVLA~\cite{kim2024openvla} and $\pi_0$~\cite{black2024pi_0}, have demonstrated strong generalization through pretraining on large and diverse robot datasets, adapting them to new tasks often still requires task-specific fine-tuning or retraining, limiting their ability to behave as truly generalist agents. In contrast, humans can rapidly adapt to novel tasks by drawing on prior experience~\cite{baddeley2006working,robertson2012new,tulving1972episodic}. Motivated by this capability, recent work has explored \emph{in-context imitation learning}, in which demonstrations of a new task are provided directly to the VLA at inference time, enabling adaptation without updating the model parameters. ICRT~\cite{fu2024context} is an early representative approach, formulating robot control as next-token prediction over sequences of visual observations, proprioceptive states, and actions. By placing one or more human teleoperation trajectories in the transformer's context, ICRT can infer and execute previously unseen tasks without test-time parameter updates, demonstrating that the in-context learning behavior of autoregressive models can extend from language to continuous robotic control. More recently, RICL~\cite{sridhar2025ricl} showed that such capability does not naturally emerge in pretrained VLAs and proposed explicitly post-training a pretrained \(\pi_0\)-FAST policy to exploit retrieved demonstrations. At deployment, RICL retrieves relevant states and actions from a small demonstration buffer and incorporates them into the model context, enabling adaptation to new manipulation tasks using roughly 10--20 demonstrations without updating model parameters. 

A parallel direction investigates whether existing frozen VLAs can benefit from demonstrations without additional in-context learning (ICL) training. Retrieval-VLA~\cite{zhang2026retrieval} enables training-free adaptation by identifying relevant past experience and directly incorporates the associated information into the prediction of a frozen VLA. Since the retrieved demonstrations are not provided as context that the VLA itself must learn to interpret, this approach avoids ICL-specific post-training. Later work further enriches the information provided by demonstrations. For example, StellaVLA~\cite{xu2026stellavla} converts retrieved trajectories into structured task plans, subgoals, and motion descriptions before conditioning the VLA on them, rather than relying only on raw trajectory similarity. Collectively, these studies establish demonstrations and retrieval as a promising interface for rapidly adapting VLAs without task-specific parameter updates.

\section{Additional Implementation Details}
\label{app:implement}

\paragraph{Evaluation Benchmark.}
We select 25 evaluation tasks from LIBERO with task IDs ranging from 20 to 69, as listed in Table~\ref{tab:libero-tasks}. Repeated instructions correspond to distinct LIBERO task IDs and scenes. Since tasks 0--19 are used to post-train the VLA model, we exclude them from evaluation to ensure that all evaluated tasks are unseen during post-training and therefore measure the test-time adaptation enabled by retrieval. We further exclude tasks on which the fixed VLA exhibits no baseline competence: for each candidate task, we run the VLA ten times using the original RICL retrieval setting and retain the task if it succeeds at least once. This avoids tasks that are beyond the capability of the current VLA, where changing the retrieval mechanism is unlikely to produce a measurable difference. Consequently, our evaluation focuses on tasks for which the VLA has at least minimal capability and where the effect of retrieval design can be meaningfully compared.

\paragraph{Model Architecture.}
We evaluate all retrieval methods using a fixed \(\pi_0\)-FAST VLA model~\cite{pertsch2025fast}, following the RICL setting. Since the original RICL model is trained on a different benchmark, we further adapt the VLA to LIBERO. Specifically, we initialize the model from Physical Intelligence's official \texttt{pi0\_fast\_libero} checkpoint~\footnote{\url{https://huggingface.co/Fisher-Wang/pi0-fast-libero-jax}} and post-train it for 10,000 steps on the first 20 LIBERO tasks (IDs 0--19), using 20 demonstrations per task. This training setup follows the same 10k-step post-training protocol used in the original RICL work~\cite{sridhar2025ricl}.

\paragraph{Hardware.}
All experiments are run on Triton, the high-performance computing cluster at Aalto University. We fine-tune the \(\pi_0\)-FAST VLA on LIBERO using two NVIDIA H100 GPUs. All retrieval methods are evaluated on a single NVIDIA GeForce RTX 5090 GPU with 32 GB of memory.

\paragraph{Retrieval Methods}
\texttt{RICL-ImageState}: This method augments the original image-based retrieval method \texttt{RICL-Image} with the robot's proprioceptive state $s$. Specifically, let \(s_t\in\mathbb{R}^8\) denote the state of the current query and \(s_i\in\mathbb{R}^8\) the state associated with a demonstration entry \(m_i\in\mathcal D\) from the retrieval buffer, where each state vector consists of 8 dimensions. We first normalize each state dimension using the mean and standard deviation computed from the LIBERO data used to post-train the VLA. Specifically, letting \(\mu_k\) and \(\sigma_k\) denote the mean and standard deviation of state for dimension \(k\), we define the normalized state \(\tilde s^{(k)}=(s^{(k)}-\mu_k)/\sigma_k\), for \(k=1,\ldots,8\). The state distance between the current query and entry \(m_i\) is then \(d_t^{\mathrm{state}}(i)=\|\tilde s_t-\tilde s_i\|_2\). We retain the original RICL image distance \(d_t^{\mathrm{img}}(i)=\|f_{\mathrm{DINO}}(I_t^{\mathrm{top}})-f_{\mathrm{DINO}}(I_i^{\mathrm{top}})\|_2\). Since the image and state distances can have different numerical scales, for each query we independently rescale the two distance terms by their median values over the retrieval buffer. Specifically, we define
\(
m_t^{\mathrm{img}}
=
\operatorname{median}_{j\in\{1,\ldots,N\}}
d_t^{\mathrm{img}}(j),
m_t^{\mathrm{state}}
=
\operatorname{median}_{j\in\{1,\ldots,N\}}
d_t^{\mathrm{state}}(j).
\)
The normalized distances are then \(
\bar d_t^{\mathrm{img}}(i)
=
\frac{d_t^{\mathrm{img}}(i)}{m_t^{\mathrm{img}}},
\bar d_t^{\mathrm{state}}(i)
=
\frac{d_t^{\mathrm{state}}(i)}{m_t^{\mathrm{state}}}.
\) We combine the two terms to obtain the final retrieval score
\[
d_t^{\mathrm{ImageState}}(i)
=
\bar d_t^{\mathrm{img}}(i)
+
\bar d_t^{\mathrm{state}}(i).
\]
The \(M\) entries with the smallest \(d_t^{\mathrm{ImageState}}(i)\) are retrieved as the in-context demonstrations.

\begin{table}[t]
\centering
\caption{The 25 LIBERO tasks used in our evaluation.}
\label{tab:libero-tasks}
\scriptsize
\setlength{\tabcolsep}{3.2pt}
\renewcommand{\arraystretch}{0.95}
\begin{tabular}{clcl}
\toprule
\textbf{ID} & \textbf{Task Instruction} & \textbf{ID} & \textbf{Task Instruction} \\
\midrule
20 & turn on the stove
& 49 & pick up the tomato sauce and put it in the basket \\

21 & turn on the stove and put the frying pan on it
& 50 & pick up the alphabet soup and put it in the basket \\

24 & put the black bowl in the bottom drawer of the cabinet
& 51 & pick up the butter and put it in the basket \\

25 & put the black bowl on top of the cabinet
& 52 & pick up the milk and put it in the basket \\

28 & close the top drawer of the cabinet
& 55 & pick up the alphabet soup and put it in the tray \\

29 & put the black bowl in the top drawer of the cabinet
& 56 & pick up the butter and put it in the tray \\

30 & put the black bowl on the plate
& 57 & pick up the cream cheese and put it in the tray \\

31 & put the black bowl on top of the cabinet
& 59 & pick up the tomato sauce and put it in the tray \\

34 & put the yellow and white mug to the front of the white mug
& 60 & pick up the black bowl on the left and put it in the tray \\

35 & open the microwave
& 61 & pick up the chocolate pudding and put it in the tray \\

43 & put the white bowl on top of the cabinet
& 68 & put the yellow and white mug on the right plate \\

44 & turn on the stove
& 69 & put the chocolate pudding to the left of the plate \\

46 & pick up the alphabet soup and put it in the basket
& & \\
\bottomrule
\end{tabular}
\end{table}

\texttt{RICL-VLAFeat}: This method investigates whether features extracted directly from the VLA model provide a more effective retrieval representation than the generic visual features produced by DINOv2. We use Physical Intelligence's \texttt{pi0\_fast\_libero} checkpoint as the feature extractor. Given the top-view image \(I_t^{\mathrm{top}}\) and the normalized proprioceptive state \(\tilde{s}_t\), the inputs are first encoded into the token representation used by \(\pi_0\)-FAST and processed by its transformer backbone. Rather than using the final action prediction, we extract an intermediate representation from the first transformer block. Specifically, we take the output-projected self-attention representation corresponding to the last valid input token, which aggregates information from the input observation before subsequent transformer processing. This produces a 2,048-dimensional feature vector, which is further \(L_2\)-normalized. We denote the resulting feature extractor by \(f_{\pi_0}\). For each demonstration entry \(m_i\in\mathcal D\), we precompute its feature as \(z_i=f_{\pi_0}(I_i^{\mathrm{top}},\tilde{s}_i)\). At timestep \(t\), the query feature is similarly computed as \(z_t=f_{\pi_0}(I_t^{\mathrm{top}},\tilde{s}_t)\). Retrieval is then performed using the Euclidean distance
\[
d_t^{\mathrm{VLAFeat}}(i)=\|z_t-z_i\|_2,
\]
and the \(M\) entries with the smallest distances are selected as the retrieved context.

For all retrieval methods, we omit the additional distance-weighted interpolation step used in the original RICL framework~\cite{sridhar2025ricl}, since its interpolation weight is defined based on the DINO retrieval distance. To ensure a fair comparison across different retrieval representations and avoid introducing a DINO-specific advantage, we exclude this interpolation step for all methods.

\paragraph{Task-hierarchical bootstrap confidence interval.}
To quantify uncertainty in the overall task success rate, we use a task-hierarchical bootstrap procedure that treats tasks, rather than individual evaluation trials, as the primary sampling units. Each of the 25 evaluation tasks is tested on 10 fixed initial scenes. For a given retrieval method, we first compute the success rate separately for each task by averaging the binary success outcomes over its 10 evaluation scenes. This produces 25 task-level success rates. We then generate one bootstrap sample by sampling 25 tasks with replacement from these 25 task-level results. For the sampled tasks, we compute the mean of their task-level success rates to obtain one bootstrapped estimate of the overall success rate. We repeat this procedure for 10,000 bootstrap samples, resulting in an empirical distribution of the overall task success rate. The \emph{task-hier. CI} is defined by the 2.5th and 97.5th percentiles of this bootstrap distribution, corresponding to a 95\% confidence interval.

\paragraph{McNemar test.}
To compare two retrieval methods under the same evaluation conditions, we use the exact McNemar test~\cite{mcnemar1947note,fagerland2013mcnemar}. Since all methods are evaluated on the same tasks and the same fixed initial scenes, their success and failure outcomes can be compared pairwise. For two methods \(A\) and \(B\), let \(n_{10}\) denote the number of evaluations for which \(A\) succeeds while \(B\) fails, and let \(n_{01}\) denote the number for which \(A\) fails while \(B\) succeeds. Evaluations on which both methods succeed or both fail do not provide evidence about their relative performance and are therefore not used by the McNemar test. Under the null hypothesis that the two methods have the same success rate, the two disagreement cases should be equally likely: \(A\) succeeds while \(B\) fails, or \(B\) succeeds while \(A\) fails. Let \(n_d=n_{10}+n_{01}\) be the total number of disagreements. Conditioning on these \(n_d\) disagreement cases, the number of cases favoring method \(A\) follows a binomial distribution,
\[
N_{10}\sim\mathrm{Binomial}(n_d,1/2),
\]
under the null hypothesis. The two-sided exact McNemar \(p\)-value is therefore
\[
p
=
2\min\left\{
\Pr(N_{10}\le n_{10}),
\Pr(N_{10}\ge n_{10})
\right\},
\]
clipped at \(1\). A small \(p\)-value indicates that the difference between \(n_{10}\) and \(n_{01}\) is statistically significant, providing evidence that methods \(A\) and \(B\) have different success rates.

\paragraph{Retrieval-quality analysis.}

To understand how retrieval quality affects the actual behavior of the VLA, we first measure the exact-task retrieval rate under two different query settings. Specifically, for a given query, the exact-task retrieval rate is defined as the fraction of the \(M\) retrieved entries whose task identity matches that of the query task. This metric reflects how often the retriever returns demonstrations from the same task, under the assumption that same-task demonstrations are more likely to provide directly relevant information for executing the target task. We first consider an \emph{offline} setting. For each of the 10 fixed evaluation scenes per task, we use states sampled from the corresponding successful expert demonstration as retrieval queries and compute the exact-task retrieval rate. We then consider a matched \emph{on-policy} setting using the same initial scenes, but query the retrieval buffer with the states actually visited by the VLA during execution. Since the expert and on-policy queries originate from the same initial scenes, this comparison isolates how retrieval behavior changes when moving from successful expert trajectories to the states encountered by the policy itself. We do not evaluate the offline setting for \texttt{RICL-Random}, since it samples entries uniformly from the retrieval buffer and therefore has an expected exact-task retrieval rate of \(1/25=4\%\).

In addition, we use Normalized Discounted Cumulative Gain (NDCG) to evaluate retrieval quality in an offline setting, where query states are sampled from held-out expert demonstrations that are not included in the retrieval buffer. The intuition is that a retrieved frame should be considered more relevant if it comes from the same task as the query and if its subsequent action chunk is similar to that of the query state. Specifically, for each query, we compare its subsequent action chunk with the action chunk associated with each candidate frame in the retrieval buffer. Each action chunk contains the next 10 actions, with each action represented by a 7-dimensional vector. Before computing their difference, each action dimension is normalized to reduce the effect of different numerical scales across action dimensions. Let \(\tilde A_q\in\mathbb{R}^{10\times 7}\) and \(\tilde A_i\in\mathbb{R}^{10\times 7}\) denote the normalized future-action chunks of the query and a candidate entry, respectively. We measure their action discrepancy using the root mean squared error
\[
e(q,i)
=
\sqrt{
\frac{1}{70}
\sum_{\tau=1}^{10}
\sum_{k=1}^{7}
\left(
\tilde A_q^{(\tau,k)}
-
\tilde A_i^{(\tau,k)}
\right)^2
}.
\]
A smaller \(e(q,i)\) indicates that the candidate frame is followed by an action sequence that is more similar to the future action sequence of the query state.

Given the action discrepancy \(e(q,i)\), we next assign a relevance score to each candidate frame in the retrieval buffer. Candidate frames from a different task than the query are assigned zero relevance. For candidate frames from the same task, we rank their action discrepancies from smallest to largest. Let \(r_i\in\{1,\ldots,N_q\}\) denote the rank of candidate \(i\) among all \(N_q\) same-task candidates, where \(r_i=1\) corresponds to the smallest action discrepancy. We define its relevance as
\[
\mathrm{rel}_i
=
1-\frac{r_i-1}{N_q-1}.
\]
Thus, same-task candidates with more similar future action chunks receive larger relevance scores, while candidates from different tasks receive zero relevance. Let \(i_1,\ldots,i_4\) denote the four retrieved entries ordered by their retrieval score. We compute
\[
\mathrm{DCG}
=
\sum_{j=1}^{4}
\frac{\mathrm{rel}_{i_j}}{\log_2(j+1)},
\]
so that highly relevant entries receive more credit when they appear earlier in the retrieved ranking. We then define
\[
\mathrm{NDCG}
=
\frac{\mathrm{DCG}}{\mathrm{IDCG}},
\]
where \(\mathrm{IDCG}\) is the DCG obtained by the ideal ranking consisting of the four same-task candidate frames with the smallest future-action discrepancy. A larger \(\mathrm{NDCG}\) therefore indicates that the retrieved top-four entries are closer to this ideal ranking in terms of both task identity and future-action similarity.

\section{Additional Experimental Results}
\label{app:result}

\begin{figure}[t]
\centering
\begin{minipage}[t]{0.24\linewidth}
    \centering
    \includegraphics[width=\linewidth]{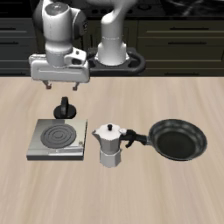}
    \vspace{-1mm}
    
\end{minipage}
\hfill
\begin{minipage}[t]{0.24\linewidth}
    \centering
    \includegraphics[width=\linewidth]{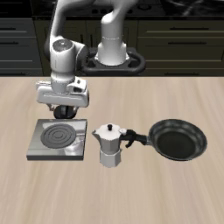}
    \vspace{-1mm}
    
\end{minipage}
\hfill
\begin{minipage}[t]{0.24\linewidth}
    \centering
    \includegraphics[width=\linewidth]{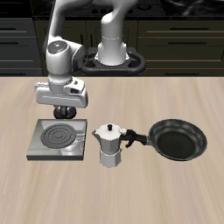}
    \vspace{-1mm}
    
\end{minipage}
\hfill
\begin{minipage}[t]{0.24\linewidth}
    \centering
    \includegraphics[width=\linewidth]{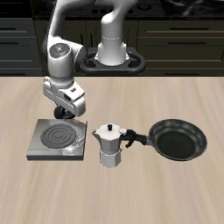}
    \vspace{-1mm}
    
\end{minipage}

\caption{A successful \texttt{RICL-Image} example on Task~20 (turn on the stove). The dominant retrieved source is Task~21 (turn on the stove and put the frying pan on it).}
\vspace{-2ex}
\label{fig:cross-task-example-t20}
\end{figure}

\begin{figure*}[t]
\centering
\scriptsize

\textbf{Task 24: put the black bowl in the bottom drawer of the cabinet}\\[1mm]

\begin{minipage}[t]{0.23\linewidth}
    \centering
    \includegraphics[width=\linewidth]{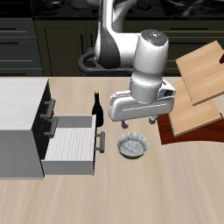}
\end{minipage}
\hfill
\begin{minipage}[t]{0.23\linewidth}
    \centering
    \includegraphics[width=\linewidth]{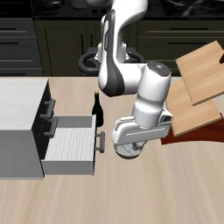}
\end{minipage}
\hfill
\begin{minipage}[t]{0.23\linewidth}
    \centering
    \includegraphics[width=\linewidth]{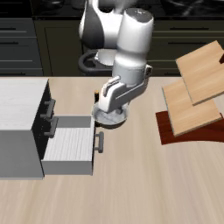}
\end{minipage}
\hfill
\begin{minipage}[t]{0.23\linewidth}
    \centering
    \includegraphics[width=\linewidth]{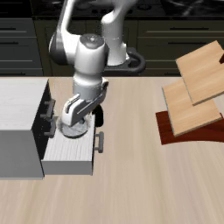}
\end{minipage}

\vspace{2mm}

\textbf{Task 25: put the black bowl on top of the cabinet}\\[1mm]

\begin{minipage}[t]{0.23\linewidth}
    \centering
    \includegraphics[width=\linewidth]{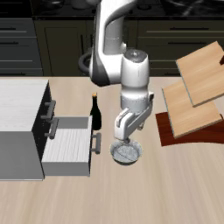}
\end{minipage}
\hfill
\begin{minipage}[t]{0.23\linewidth}
    \centering
    \includegraphics[width=\linewidth]{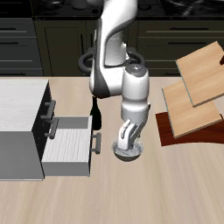}
\end{minipage}
\hfill
\begin{minipage}[t]{0.23\linewidth}
    \centering
    \includegraphics[width=\linewidth]{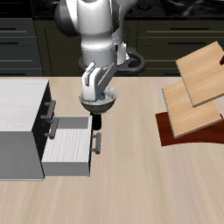}
\end{minipage}
\hfill
\begin{minipage}[t]{0.23\linewidth}
    \centering
    \includegraphics[width=\linewidth]{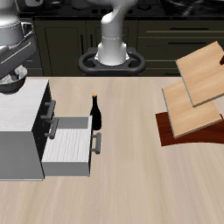}
\end{minipage}

\caption{Successful examples on Task~24 and Task~25, where each task uses the other as the dominant retrieval source. The two tasks involve the same object and highly similar scene structures, but require different target locations.}
\label{fig:t24-t25-comparison}
\end{figure*}

As shown in Figure~\ref{fig:cross-task-example-t20}, when performing Task~20 (turn on the stove), \texttt{RICL-Image} repeatedly retrieves demonstrations from the related Task~21 (turn on the stove and put the frying pan on it), yet still achieves (100\%) success. These examples demonstrate that cross-task demonstrations can provide useful transferable information when the retrieved task shares relevant manipulation knowledge with the target task.

We also observe systematic retrieval between tasks with similar scene structure but different goals, such as Task~24 (put the black bowl in the bottom drawer of the cabinet) and Task~25 (put the black bowl on top of the cabinet), as shown in Figure~\ref{fig:t24-t25-comparison}. Although these two tasks involve the same object and environment, they require different target locations. This suggests that retrieval from a different task should not be automatically regarded as incorrect. Its usefulness depends on whether the retrieved demonstration provides action-relevant information consistent with the current task goal.

\end{document}